\documentclass[twoside]{article}

\usepackage[accepted]{aistats2020}

\usepackage[square]{natbib}

\usepackage{amsmath, amsthm, amsfonts}

\usepackage{subcaption}
\usepackage{pdfpages}
\usepackage{makecell}
\usepackage{todonotes}
\usepackage{color}
\usepackage{caption}
\usepackage{tabularx}
\usepackage{graphicx}
\usepackage{url}
\usepackage{hyperref}
\usepackage{relsize}
\usepackage{algorithm} 
\usepackage{algpseudocode}
\usepackage{mathtools}
\usepackage[percent]{overpic}

\usepackage{color}

\DeclareMathOperator*{\argmax}{\arg\!\max}
\DeclareMathOperator*{\argmin}{arg\,min} 

\algtext*{EndWhile}
\algtext*{EndIf}
\algtext*{EndFor}

\begin{document}

\twocolumn[
\aistatstitle{Coping With Simulators That Don't Always Return}

\aistatsauthor{ Andrew Warrington \And Saeid Naderiparizi \And  Frank Wood }

\aistatsaddress{University of Oxford \\ \url{andreww@robots.ox.ac.uk} \And  University of British Columbia \\ \url{saeidnp@cs.ubc.ca} \And University of British Columbia \\ \url{fwood@cs.ubc.ca}}
]

\begin{abstract}
Deterministic models are approximations of reality that are easy to interpret and often easier to build than stochastic alternatives.  
Unfortunately, as nature is capricious, observational data can never be fully explained by deterministic models in practice.  
Observation and process noise need to be added to adapt deterministic models to behave stochastically, such that they are capable of explaining and extrapolating from noisy data.
We investigate and address computational inefficiencies that arise from adding process noise to deterministic simulators that fail to return for certain inputs; a property we describe as ``brittle.''
We show how to train a conditional normalizing flow to propose perturbations such that the simulator succeeds with high probability, increasing computational efficiency.
\end{abstract}

\section{Introduction}

In order to compensate for epistemic uncertainty due to modelling approximations and unmodelled aleatoric uncertainty, deterministic simulators are often ``converted'' to ``stochastic'' simulators by perturbing the state at each time step.
In practice this allows the simulator to explain the variability observed in real data without requiring excessive observation noise.
Such models are more resilient to misspecification, are capable of providing uncertainty estimates, and provide better inferences in general~\citep{moller2011parameter, Saarinen2008hh, LV200874, stochchem, renard2013stochastic}.

Often, state-independent Gaussian noise with heuristically tuned variance is used to perturb the state~\citep{adhikari2013introductory, brockwell2016introduction, fox1997stochastic, reddy2016simulating, du2006dynamics, allen2017primer, Mbalawata2013}.
However, naively adding noise to the state will, in many applications, render the perturbed input state ``invalid,''
where invalid states cause the simulator to raise an exception and not return a value~\citep{RAZAVI201995, lucas2013failure, gmd2019crash}. 
We formally define failure by extending the possible output of the simulator to include $\bot$ (read as ``bottom'') denoting simulator failure.
The principal contribution of this paper is a technique for avoiding invalid states by choosing perturbations that minimize the failure rate.
The technique we develop results in a reduction in simulator failures, while maintaining the original model.

Examples of failure modes include ordinary differential equation (ODE) solvers not converging to the required tolerance in the allocated time, or, the perturbed state entering into an unhandled configuration, such as solid bodies intersecting.
Establishing the state-perturbation pairs that cause failure is non-trivial.
Hence, the simulation artifact can be sensitive to seemingly inconsequential alterations to the state -- a property we describe as ``brittle.'' 
Failures waste computational resources and reduce the diversity of simulations for a finite sample budget, for instance, when used as the proposal in sequential Monte Carlo.
As such, we wish to learn a proposal over perturbations such that the simulator exits with high probability, but renders the joint distribution unchanged.

We proceed by framing sampling from brittle simulators as rejection samplers, then seek to eliminate rejections by estimating the state-dependent density over perturbations that do not induce failure.
We then demonstrate that using this learned proposal yields lower variance results when used in posterior inference with a fixed sample budget, such as pseudo-marginal evidence estimates produced by sequential Monte Carlo sweeps.
Source code for reproduction of figures and results in this paper is available at \url{https://github.com/plai-group/stdr}.

\section{Background}
\subsection{Smoothing Deterministic Models}
Deterministic simulators are often stochastically perturbed to increase the diversity of the achievable simulations and to fit data more effectively.
The most widespread example of this is perturbing linear dynamical systems with Gaussian noise at each timestep.
The design of the system is such that the distribution over state at each point in time is Gaussian distributed.
However, the simplistic dynamics of such a system may be insufficient for simulating more complex systems.
Examples of such systems are: stochastic models of neural dynamics~\citep{fox1997stochastic, coutin2018fractional, goldwyn2011hh, Saarinen2008hh}, econometrics~\citep{lopes2011finance}, epidemiology~\citep{allen2017primer} and mobile robotics~\citep{thrun2001robust, fallon2012efficient}.
In these examples, the simulator state is perturbed with noise drawn from a distribution and is iterated using the simulator to create discrete approximations of the distribution over state as a function of time.

\subsection{Simulator Failure}
As simulators become more complex, guaranteeing the simulator will not fail for perturbed inputs becomes more difficult, and individual function evaluations become more expensive.
\citet{lucas2013failure} and \citet{edwards2011precalibrating} establish the sensitivity of earth science models to global parameter values by building a discriminative classifier for parameters that induce failure. 
\citet{gmd2019crash} take an alternative approach instead treating simulator failure as an imputation problem, fitting a function regressor to predict the outcome of the failed experiment given the neighboring experiments that successfully terminated.
However these methods are limited by the lack of clear probabilistic interpretation in terms of the originally specified joint distribution in time series models, their ability to scale to high dimensions, and their applicability to state-space models.

\subsection{State-space Inference and Model Selection}
Probabilistic models are ultimately deployed to make inferences about the world.
Hence the goal is to be able to recover distributions over unobserved states, predict future states and learn unknown parameters of the model from data.
Posterior state-space inference refers to the task of recovering the distribution $p_{\mathcal{M}}(\mathbf{x}_{0:T} | \mathbf{y}_{1:T})$, where $\mathbf{x}_{0:T}$ are the latent states, $\mathbf{y}_{1:T}$ are the observed data, and $\mathcal{M}$ denotes the model if multiple different models are available.
Inference in Gaussian perturbed linear dynamical systems can be performed using techniques such as Kalman smoothing~\citep{kalman1960new}, however, the restrictions on such techniques limit their applicability to complex simulators, and so numerical methods are often used in practice. 

\begin{algorithm}[t]
 \caption{Sequential Monte Carlo}\label{alg:meth:smc_p}
 \begin{algorithmic}[1]
  \Procedure{SMC}{$p_{\mathcal{M}}(\mathbf{x}_0)$, $\overline{p}_{\mathcal{M}}(\mathbf{x}_t | \mathbf{x}_{t-1})$, $\mathbf{y}_{1:T}$, $p_{\mathcal{M}}(\mathbf{y}_t | \mathbf{x}_t)$, $N$} 
    \For{$n=1:N$}
        \State $\mathbf{x}_0^{(n)} \sim p_{\mathcal{M}}(\mathbf{x}_0)$ \Comment{Initialize from prior.}
    \EndFor
    \State $L_{\mathcal{M}} \gets 0$ \Comment{Track log-evidence} 
     \For{$t=1:T$}
        \For{$n=1:N$}
          \State $\tilde{\mathbf{x}}_t^{(n)} \sim \overline{p}_{\mathcal{M}}\left(\mathbf{x}_t | \mathbf{x}_{t-1}^{(n)}\right)$  \label{alg:meth:smc_p:p} \Comment{Alg \ref{alg:rs}.}
          \State $w^{(n)}_t \gets p_{\mathcal{M}}\left(\mathbf{y}_t | \tilde{\mathbf{x}}_t^{(n)}\right)$ \Comment{Score particle.} \label{alg:meth:smc_p:w}
        \EndFor
        \For{$n=1:N$} \Comment{Normalize weights.}
          \State $W^{(n)}_t \gets w^{(n)}_t / \sum_{i=1}^N w^{(i)}_t$ 
        \EndFor
        \For{$n=1:N$} \Comment{Apply resampling.}
          \State $a^{(n)}_t \sim \text{Discrete}\left(\mathbf{W}_t\right)$ \label{alg:meth:smc_p:a}
          \State $\mathbf{x}_{t}^{(n)} \gets \tilde{\mathbf{x}}_{t}^{\left(a^{(n)}_t\right)}$ 
        \EndFor
        \State $L_{\mathcal{M}} \gets L_{\mathcal{M}} + \log\left(\frac{1}{N}\sum_{i=1}^N w^{(i)}_t\right)$ \label{alg:meth:smc_p:e}
     \EndFor
     \State $\textbf{return}\ \mathbf{x}^{(1:N)}_{0:T},\ a_{1:T}^{(1:N)},\ L_{\mathcal{M}}$ 
  \EndProcedure
 \end{algorithmic}
\end{algorithm}

A common method for performing inference in complex, simulation based models is sequential Monte Carlo (SMC)~\citep{doucet2001introduction}.
The basic algorithm for SMC is shown in Algorithm \ref{alg:meth:smc_p}, where $p_{\mathcal{M}}(\mathbf{x}_0)$ is the prior over initial state, $\overline{p}_{\mathcal{M}}(\mathbf{x}_t | \mathbf{x}_{t-1})$ is the dynamics model, or simulator, $p_{\mathcal{M}}(\mathbf{y}_t | \mathbf{x}_t)$ is the likelihood, defining the relationship between latent states and observed data, and $N$ is the number of particles used.
On a high level, SMC produces a discrete approximation of the target distribution by iterating particles through the simulator, and then preferentially continuing those simulations that ``explain'' the observed data well.
While a detailed understanding of particle filtering is not required, the core observation required for this work is that the likelihood of failed simulations is defined as zero: $p(\mathbf{y}_t | \mathbf{x}_t = \bot) \coloneqq 0$, and hence are rejected with certainty.

Posterior inference pipelines often also provide estimates of the model evidence, $p_{\mathcal{M}}(\mathbf{y}_{1:T})$.
SMC provides such an estimate, referred to as a pseudo-marginal evidence, denoted in Algorithm~\ref{alg:meth:smc_p} as $L_{\mathcal{M}}$.
This pseudo-marginal evidence is calculated (in log space) as the sum of the expected value of the unnormalized importance weights (Algorithm \ref{alg:meth:smc_p}, Lines \ref{alg:meth:smc_p:w} and \ref{alg:meth:smc_p:e}).
This evidence can be combined with the prior probability of each model via Bayes rule to estimate the posterior probability of the model (up to a normalizing constant)~\citep{mackay2003information}.
These posteriors can be compared to perform Bayesian model selection, where the model with the highest posterior is selected and used to perform inference.
This is often referred to as marginal maximum \emph{a posteriori} parameter estimation (or model selection)~\citep{doucet2002marginal, kantas2015particle}.
Recent work investigates model selection using approximate, likelihood-free inference techniques~\citep{papamakarios2019sequential, lueckmann2019likelihood}, however, we do not consider these methods here, instead focusing on mitigating computational inefficiencies arising directly from simulator failure.

\section{Methodology}
\label{sec:meth}

We consider deterministic models, expressed as simulators, describing the time-evolution of a state $\mathbf{x}_t \in \mathcal{X}$, where we denote application of the simulator iterating the state as $\mathbf{x}_t \gets f(\mathbf{x}_{t-1})$.
A stochastic, additive perturbation to state, denoted $\mathbf{z}_t \in \mathcal{X}$, is applied to induce a distribution over states.
The distribution from which this perturbation is sampled is denoted $p(\mathbf{z}_t | \mathbf{x}_{t-1})$, although, in practice, this distribution is often state independent.
The iterated state is then calculated as $\mathbf{x}_t \gets f(\mathbf{x}_{t-1} + \mathbf{z}_t)$.

However, we consider the case where the simulator can fail for ``invalid'' inputs, denoted by a return value of $\bot$.
Hence the complete definition of $f$ is ${f:\mathcal{X} \rightarrow \left\lbrace \mathcal{X}, \bot \right\rbrace}$.
The region of valid inputs is denoted as ${\mathcal{X}_{A} \subset \mathcal{X}}$, and the region of invalid inputs as ${\mathcal{X}_{R} \subset \mathcal{X}}$, such that ${\mathcal{X}_{A} \sqcup \mathcal{X}_{R} = \mathcal{X}}$, where the boundary between these regions is unknown.
Over the whole support, $f$ defines a many-to-one function, as $\mathcal{X}_R$ maps to $\bot$.
However, the algorithm we derive only requires that $f$ is one-to-one in the accepted region.
This is not uncommon in real simulators, and is satisfied by, for example, ODE models.
We define the random variable $A_t \in \left\lbrace 0, 1 \right\rbrace$ to denote whether the state-perturbation pair does not yield simulator failure and is ``accepted.''

We define the iteration of perturbed deterministic simulator as a rejection sampler, with a well-defined target distribution (\S\ref{sec:meth:rs}).
We use this definition and the assumptions on $f$ to show that we can target the same distribution by learning the state-conditional density of perturbations, conditioned on acceptance (\S\ref{sec:meth:cov}).
We train an autoregressive flow to fit this density (\S\ref{sec:meth:training}), and describe how this can be used in inference, highlighting the ease with which it can be inserted into a particle filter (\S\ref{sec:meth:using-q}).
We empirically show that using this learned proposal distribution in place of the original proposal improves the performance of particle-based state-space inference methods (\S\ref{sec:experiments}).

\subsection{Brittle Simulators as Rejection Samplers}
\label{sec:meth:rs}
The naive approach to sampling from the perturbed system, shown in Algorithm \ref{alg:rs}, is to repeatedly sample from the proposal distribution and evaluate $f$ until the simulator successfully exits.
This procedure defines $A_t = \mathbb{I} \left[ f(\mathbf{x}_{t-1} + \mathbf{z}_t) \neq \bot \right],\ \mathbf{z}_t \sim p(\mathbf{z}_t | \mathbf{x}_{t-1})$, i.e. successfully iterated samples are accepted with certainty.
This incurs significant wasted computation as the simulator must be called repeatedly, with failed iterations being discarded.
The objective of this work is to derive a more efficient sampling mechanism. 

\begin{figure}[t]
    \centering
    \includegraphics[width=0.45\textwidth]{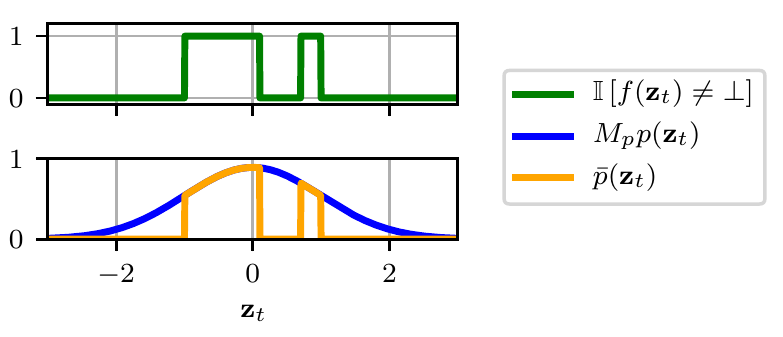}
    \vspace{-0.3cm}
    \caption{Graphical representation of how a brittle deterministic simulator acts as a rejection sampler, targeting $\overline{p}(\mathbf{z}_t | \mathbf{x}_{t-1})$.
    We set $\mathbf{x}_t = 0$ for clarity.
    The simulator, $f(\mathbf{z}_t)$, returns $\bot$ for unknown input regions, shown in green.
    The proposal over $\mathbf{z}_t$ is shown in blue.
    The target distribution, $\overline{p}(\mathbf{z}_t)$, shown in orange, is implicitly defined as $\overline{p}(\mathbf{z}_t) = \frac{1}{M_p} p(\mathbf{z}_t) \mathbb{I}\left[ f(\mathbf{z}_t) \neq \bot \right]$, where $M_p$ is the normalizing constant from $\overline{p}$, equal to the acceptance rate.
    Accordingly, the proposal distribution, scaled by $M_p$, is \emph{exactly} equal to $\overline{p}(\mathbf{z}_t)$ in the accepted region.
    Algorithm \ref{alg:rs} therefore implicitly constructs a rejection sampler, where the acceptance criterion reduces to $\mathbb{I}\left[ f(\mathbf{z}_t) \neq \bot \right]$, without needing to specify any additional scaling constants.
    }
    \label{fig:rs}
\end{figure}

We begin by establishing Algorithm \ref{alg:rs} as a rejection sampler, targeting the distribution over successfully iterated states.
This reasoning is illustrated in Figure~\ref{fig:rs}.
The behavior of $f$ and the distribution $p(\mathbf{z}_t | \mathbf{x}_{t-1})$ implicitly define a distribution over successfully iterated states.
We denote this ``target'' distribution as $\overline{p}(\mathbf{x}_t | \mathbf{x}_{t-1}) = p(\mathbf{x}_t | \mathbf{x}_{t-1}, A_t = 1)$, where the bar indicates that the sample was accepted, and hence places no probability mass on failures.
Note there is no bar on $p(\mathbf{z}_t | \mathbf{x}_{t-1})$, indicating that it is defined before the accept/reject behaviors of $f$ and hence probability mass may be placed on regions that yield failure.
The functional form of $\overline{p}$ is unavailable, and the density cannot be evaluated for any input value.

\begin{algorithm}[t]
 \caption{Iterate brittle simulator, $\overline{p}(\mathbf{x}_t | \mathbf{x}_{t-1})$.}\label{alg:rs}
 \begin{algorithmic}[1]
  \Procedure{IterateSimulator}{$f$, $\mathbf{x}_{t-1}$}
     \State $\mathbf{x}_t \gets \bot$
     \While{$\mathbf{x}_t == \bot$} 
      \If{$q_{\phi}\ \texttt{is trained}$}
      \State $\mathbf{z}_t \sim q_{\phi}(\mathbf{z}_t | \mathbf{x}_{t-1})$ \Comment{Perturb under $q_{\phi}$.} \label{alg:meth:brittle:p}
      \Else
      \State $\mathbf{z}_t \sim p(\mathbf{z}_t | \mathbf{x}_{t-1})$ \Comment{Perturb under $p$.}
      \EndIf
      \State $\mathbf{x}_t \gets f( \mathbf{x}_{t-1} + \mathbf{z}_t)$ \Comment{Iterate simulator.}
     \EndWhile
     \State $\textbf{return}\ \mathbf{x}_t$
  \EndProcedure
 \end{algorithmic}
\end{algorithm}

The existence of $\overline{p}(\mathbf{x}_t | \mathbf{x}_{t-1})$ implies the existence of a second distribution: the distribution over \emph{accepted} perturbations, denoted $\overline{p}(\mathbf{z}_t | \mathbf{x}_{t-1})$.
Note that this distribution is also conditioned on acceptance under the chosen simulator, indicated by the presence of a bar.
We assume $f$ is one-to-one in the accepted region, and so the change of variables rule can be applied to directly relate this to $\overline{p}(\mathbf{x}_t | \mathbf{x}_{t-1})$.
Under our initial algorithm for sampling from a brittle simulator we can therefore write the following identity:
\begin{equation}
\overline{p}(\mathbf{z}_t | \mathbf{x}_{t-1}) = \begin{cases}
      \frac{1}{M_p} p(\mathbf{z}_t | \mathbf{x}_{t-1}), & \text{if}\ f(\mathbf{x}_{t-1} + \mathbf{z}_t)\neq \bot \label{equ:pbarprob}\\
      0, & \text{otherwise}
    \end{cases}
\end{equation}
where the normalizing constant $M_p$ is the acceptance rate under $p$.
\eqref{equ:pbarprob} indicates accepting with certainty perturbations that exit successfully can be seen as proportionally shifting mass from regions of $p$ where the simulator fails to regions where it does not. 
We exploit this definition to learn an efficient proposal.

\subsection{Change of Variable in Brittle Simulator}
\label{sec:meth:cov}
We now derive how we can learn the proposal distribution, denoted $q_{\phi}$ and parameterized by $\phi$, to replace $p$, such that the acceptance rate under $q_{\phi}$ (denoted $M_{q_{\phi}}$) tends towards unity, minimizing wasted computation.
We denote $q_{\phi}$ as the proposal we train, which, coupled with the simulator, implicitly defines a proposal over accepted samples, denoted $\overline{q}_{\phi}$.

Expressing this mathematically, we wish to minimize the distance between joint distribution implicitly specified over \emph{accepted} iterated states using the \emph{a priori} specified proposal distribution, $\overline{p}$, and $\overline{q}_{\phi}$:
\begin{equation}
    \phi^* = \argmin_{\phi} \mathbb{E}_{p(\mathbf{x}_{t-1})} \left[ \mathcal{D}_{\text{KL}} \left[ \overline{p}(\mathbf{x}_t | \mathbf{x}_{t-1}) || \overline{q}_{\phi}(\mathbf{x}_t | \mathbf{x}_{t-1}) \right] \right],\label{equ:sim:target}
\end{equation}
where we select the Kullback-Leibler (KL) divergence as the metric of distance between distributions.
The outer expectation defines this objective as amortized across state space, where we can generate the samples by directly sampling trajectories from the model~\citep{le2016inference, gershman2014amortized}.
We use the forward KL as opposed to the reverse KL, $\mathcal{D}_{\text{KL}} \left[ \overline{q}_{\phi}(\mathbf{x}_t | \mathbf{x}_{t-1}) || \overline{p}(\mathbf{x}_t | \mathbf{x}_{t-1}) \right]$, as high-variance REINFORCE estimators must be used to obtain the the differential with respect to $\phi$ of the reverse KL. 

Expanding the KL term yields:
\begin{align}
\phi^* &= \argmin_{\phi} \mathbb{E}_{p(\mathbf{x}_{t-1})} \mathbb{E}_{\overline{p}(\mathbf{x}_t | \mathbf{x}_{t-1})} \left[ \log\left( w \right) \right], \label{equ:kl_w}\\
w &= \frac{\overline{p}(\mathbf{x}_t | \mathbf{x}_{t-1})}{\overline{q}_{\phi}(\mathbf{x}_t | \mathbf{x}_{t-1})}.
\end{align}
Noting that $\overline{q}_{\phi}$ and $\overline{p}$ are defined only on accepted samples, where $f$ is one-to-one, we can apply a change of variables defined for $\overline{q}_{\phi}$ as:
\begin{equation}
   \overline{q}_{\phi}(\mathbf{x}_t | \mathbf{x}_{t-1}) = \overline{q}_{\phi} ( f^{-1} (\mathbf{x}_{t}) | \mathbf{x}_{t-1})  \left| \frac{\text{d}f^{-1} (\mathbf{x}_{t})}{\text{d}\mathbf{x}_t} \right| , \label{equ:cov_definition}
\end{equation}
and likewise for $p$.
This transforms the distribution over $\mathbf{x}_{t}$ into a distribution over $\mathbf{z}_{t}$ and a Jacobian term:
\begin{align}
w = \frac{\overline{p} (f^{-1} (\mathbf{x}_{t}) | \mathbf{x}_{t-1})  \left| \frac{\text{d}f^{-1} (\mathbf{x}_{t})}{\text{d}\mathbf{x}_t} \right|} {\overline{q}_{\phi} (f^{-1} (\mathbf{x}_{t}) | \mathbf{x}_{t-1})  \left| \frac{\text{d}f^{-1} (\mathbf{x}_{t})}{\text{d}\mathbf{x}_t} \right|}.
\end{align}
taking care to also apply the change of variables in the distribution we are sampling from in \eqref{equ:kl_w}.
Noting that the same Jacobian terms appear in the numerator and denominator we are able to cancel these:
\begin{align}
w = \frac{\overline{p} (f^{-1} (\mathbf{x}_{t}) | \mathbf{x}_{t-1}) } {\overline{q}_{\phi} (f^{-1} (\mathbf{x}_{t}) | \mathbf{x}_{t-1}) }.
\end{align}
We can now discard the $\overline{p}$ term as it is independent of $\phi$.
Noting $f^{-1}(\mathbf{x}_t) = \mathbf{x}_{t-1} + \mathbf{z}_t$ we can write \eqref{equ:sim:target} as:
\begin{align}
\phi^* &=  \argmax_{\phi} \mathbb{E}_{p(\mathbf{x}_{t-1})} \mathbb{E}_{\overline{p}(\mathbf{z}_t | \mathbf{x}_{t-1})} \hspace*{-0.1cm}\left[ \log \overline{q}_{\phi}(\mathbf{z}_t |  \mathbf{x}_{t-1}) \right]\hspace*{-0.1cm}.\label{equ:sim:p_bar_target}
\end{align}
However, this distribution is defined \emph{after} rejection sampling, and can only be defined as in \eqref{equ:pbarprob}:
\begin{align}
    \overline{q}_{\phi}(\mathbf{z}_t | \mathbf{x}_{t-1}) &= q_{\phi}(\mathbf{z}_t | \mathbf{x}_{t-1}, A_t = 1) \\%
    &=  \begin{cases}
    \frac{1}{M_{q_{\phi}}} q_{\phi}(\mathbf{z}_t | \mathbf{x}_{t-1})\ &\text{if}\ f(\mathbf{x}_{t-1} + \mathbf{z}_t) \neq \bot,\\
    0 & \text{otherwise},
    \end{cases}\nonumber
\end{align}
denoting $M_{q_{\phi}}$ as the acceptance rate under $q_{\phi}$.

However, there is an infinite family of $q_{\phi}$ proposals that yield $\overline{p} = \overline{q}_{\phi}$, each a maximizer of \eqref{equ:sim:p_bar_target} but with different rejection rates.
Noting however that there is only a single $q_{\phi}$ that has a rejection rate of zero \emph{and} renders $\overline{q}_{\phi} = \overline{p}$, and that this distribution also renders $q_{\phi} = \overline{q}_{\phi}$, we can instead optimize $q_{\phi}(\mathbf{z}_t | \mathbf{x}_{t-1})$:
\begin{equation}
\phi^* = \argmax_{\phi} \mathbb{E}_{p(\mathbf{x}_{t-1})} \mathbb{E}_{\overline{p}(\mathbf{z}_t  | \mathbf{x}_{t-1})} \left[ \log q_{\phi} ( \mathbf{z}_t | \mathbf{x}_{t-1}) \right],\label{equ:diff_target}
\end{equation}
with no consideration of rejection behavior under $q_{\phi}$.

One might alternatively try to achieve low rejection rates by adding a regularization term to \eqref{equ:sim:p_bar_target} penalizing high $M_{q_{\phi}}$. 
However, differentiation of $M_{q_{\phi}}$ is intractable, meaning direct optimization of \eqref{equ:sim:p_bar_target} is intractable.

The objective stated in \eqref{equ:diff_target} implicitly rewards $q_{\phi}$ distributions that place minimal mass on rejections by placing as much mass on accepted samples as possible.
This expressionn is differentiable with respect to $\phi$ and so we can maximize this quantity through gradient ascent with minibatches drawn from $p(\mathbf{x}_{t-1})$.
This expression shows that we can learn the distribution over accepted $\mathbf{x}_t$ values by learning the distribution over the accepted $\mathbf{z}_t$, \emph{without} needing to calculate the Jacobian or inverse of $f$.
Doing so minimizes wasted computation, targets the same overall joint distribution, and retains interpretability by utilizing the simulator.

\begin{algorithm}[t]
 \caption{Training $q_{\phi}$}\label{alg:meth:training_af}
 \begin{algorithmic}[1]
  \Procedure{TrainQ}{$p(\mathbf{x})$, $\overline{p}(\mathbf{z}_t | \mathbf{x}_{t-1})$, $K$, $N$, $q$, $\phi_0$, $\eta$}
     \For{$k=0:K-1$}
      \For{$n=1:N$}
          \State $\mathbf{x}_{t-1}^{(n)} \sim p(\mathbf{x})$ \Comment{Sample from prior.}
          \State $\mathbf{z}_t^{(n)} \sim \overline{p}(\mathbf{z}_t^{(n)} | \mathbf{x}_{t-1}^{(n)})$ \Comment{Sample noise.}
      \EndFor
      \State $E_k \gets \prod_{n=1}^N q_{\phi_k}\left(\mathbf{z}^{(n)}_{t} | \mathbf{x}^{(n)}_{t-1}\right)$
      \State $\mathbf{G}_k \gets \nabla_{\phi_k} E_k$  \Comment{Do backprop.}
      \State $\phi_{k+1} \gets \phi_k + \eta\left(\mathbf{G}_k\right)$ \Comment{Apply update.}
     \EndFor
     \State $\textbf{return}\ \phi_K$  \Comment{Return learned parameters.}
  \EndProcedure
 \end{algorithmic}
\end{algorithm}

\subsection{Training $q_{\phi}$}
\label{sec:meth:training}
To train $q_{\phi}(\mathbf{z}_t|\mathbf{x}_{t-1})$ we first define a method for sampling state-perturbation pairs.
We initialize the simulator to a state sampled from a distribution over initial value, and then iterate the perturbed simulator forward for some finite period. 
All state-perturbation pairs sampled during the trajectory are treated as a training example, and, in total, represent a discrete approximation to the prior over state for all time, and accepted state-conditional perturbation, i.e. $\mathbf{x}_{t-1} \sim p(\mathbf{x})$ and $\mathbf{z}_{t} \sim \overline{p}(\mathbf{z}_t | \mathbf{x}_{t-1})$.

We train the conditional density estimator $q_{\phi}(\mathbf{z}_t|\mathbf{x}_{t-1})$ using these samples by maximizing the conditional density of the sampled perturbation under the true distribution, as in \eqref{equ:diff_target}, as this minimizes the desired KL divergence originally stated in \eqref{equ:sim:target}. Our conditional density estimator is fully differentiable and can be trained through stochastic gradient ascent as shown in Algorithm~\ref{alg:meth:training_af}. 
The details of the chosen architecture is explained in \S\ref{sec:meth:implementation}.
The result of this procedure is an artifact that approximates the density over valid perturbations conditioned on state, $\overline{p}(\mathbf{z}_t | \mathbf{x}_{t-1})$.

\begin{figure}
    \centering
    \includegraphics[width=\columnwidth]{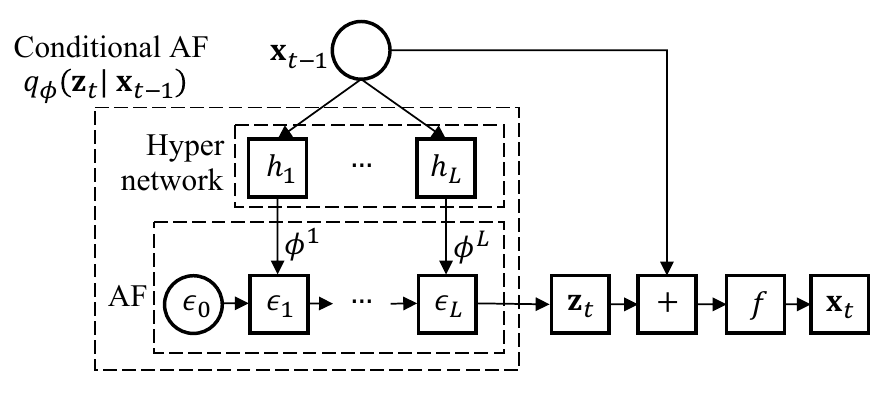}
    \caption{Diagram visualizing how $q_{\phi}$ is structured and used. 
    The previous state is input to the hypernetwork, a series of $L$ single layer neural networks, denoted $h_l$.
    Each network outputs parameters, denoted $\phi^l$, for each of the $L$ layers in the flow conditioned on the state.
    The flow samples a perturbation as $\mathbf{z}_t \sim q_{\phi}\left(\mathbf{z}_t | \mathbf{x}_{t-1} \right)$, with the internal states of the flow denoted by $\epsilon_l$.
    This perturbation is summed with the previous state and passed through the simulator, $f$, outputting the iterated state, $\mathbf{x}_t$.
    }
    \label{fig:meth:cnf_arch}
\end{figure}

\subsection{Using $q_{\phi}$}
\label{sec:meth:using-q}
Once $q_{\phi}$ has been trained, it can be deployed to enhance posterior inference, by replacing samples from $p(\mathbf{z}_t | \mathbf{x}_{t-1})$ with $q_{\phi}(\mathbf{z}_t | \mathbf{x}_{t-1})$.
We highlight here the ease with which it can be introduced into an SMC sweep.
The state is iterated by sampling from $\overline{p}(\mathbf{x}_t | \mathbf{x}_{t-1})$ on Line \ref{alg:meth:smc_p:p} of Algorithm \ref{alg:meth:smc_p}, where this sampling procedure is defined in Algorithm \ref{alg:rs}.
Instead of sampling from $p(\mathbf{z}_{t} | \mathbf{x}_{t-1})$ in Algorithm \ref{alg:rs}, the sample is drawn from $q_{\phi}(\mathbf{z}_t | \mathbf{x}_{t-1})$, and as such the sample is more likely to be accepted.
This modification requires only changing a single function call made inside the implementation of Algorithm \ref{alg:rs}.

\subsection{Implementation}
\label{sec:meth:implementation}
We parameterize the density $q_{\phi}$ using an autoregressive flow (AF)~\citep{larochelle2011neural}.
Flows define a parameterized density estimator that can be trained using stochastic gradient descent, and variants have been used in image generation~\citep{kingma2018glow}, as priors for variational autoencoders~\citep{kingma2016improved}, and in likelihood-free inference~\citep{papamakarios2019sequential,lueckmann2019likelihood}.

Specifically, we structure $q_{\phi}$ using a masked autoregressive flow~\citep{papamakarios2017masked}, with $5$ single-layer MADE blocks~\citep{germain2015made}, and batch normalization at the input to each intermediate MADE block.
The dimensionality of the flow is the number of states perturbed in the original model. 
We implement conditioning through the use of a hypernetwork~\citep{ha2016hypernetworks}, which outputs the parameters of the flow layers given $\mathbf{x}_{t-1}$ as input, as shown in Figure \ref{fig:meth:cnf_arch}. 
The hypernetworks are single-layer neural networks defined per flow layer.
Together, the flow and hypernetwork define $q_{\phi}(\mathbf{z}_t|\mathbf{x}_{t-1})$, and can be jointly trained using stochastic gradient descent. 
The networks are implemented in PyTorch~\citep{paszke2017automatic} and are optimized using ADAM~\citep{kingma2014adam}.

\section{Experiments}
\label{sec:experiments}

\subsection{Toy Problem -- Annulus}
\label{sec:annulus}
We first demonstrate our approach on a toy problem.
The true generative model of the observed data is a constant speed circular orbit around the origin in the $x$-$y$ plane, such that $\mathbf{x}_t = \left\lbrace x_t, y_t, \dot{x}_t, \dot{y}_t \right\rbrace \in  \mathbb{R}^4$.
To analyze this data we use a misspecified model that only simulates linear forward motion.
To overcome the model mismatch and fit the observed data, we add Gaussian noise to position and velocity.
We impose a failure constraint limiting the change in the distance of the point from the origin to a fixed threshold.
This condition mirrors our observation that states in brittle simulators have large allowable perturbations in particular directions, but very narrow permissible perturbations in other directions.
The true radius is unknown and so we must amortize over possible radii.

\begin{figure}[h!]
\centering

    \begin{subfigure}[t]{0.48\textwidth}
    \centering
        \includegraphics[width=0.95\textwidth]{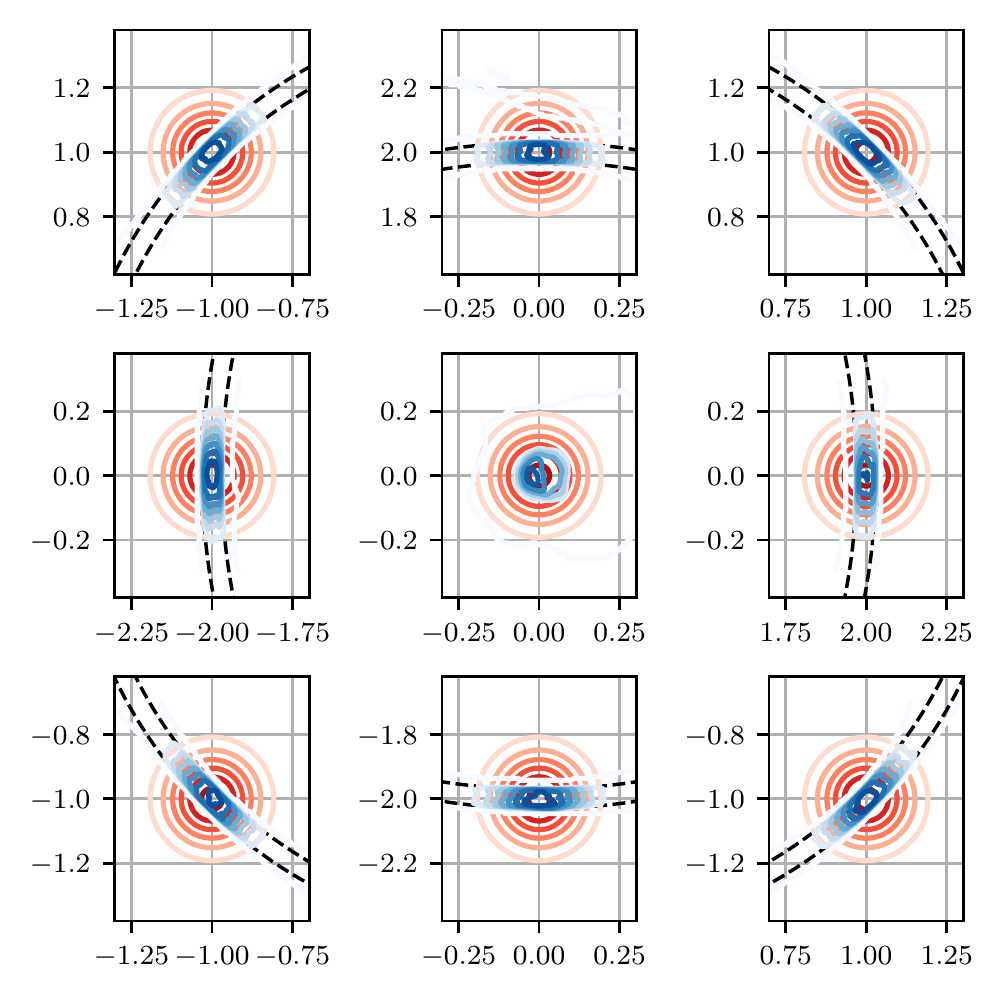}
        \vspace*{-0.2cm}
        \caption{}
        \label{fig:ring:space}
    \end{subfigure}%

    \begin{subfigure}[t]{0.24\textwidth}
        \includegraphics[width=\textwidth]{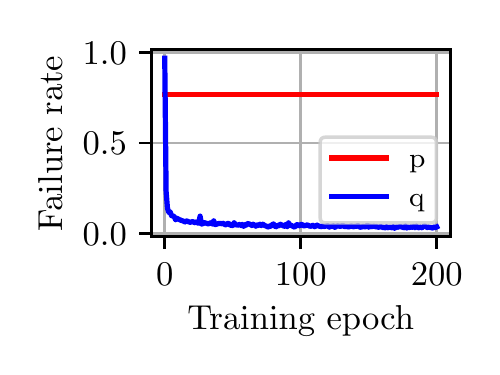}
        \vspace*{-0.7cm}
        \caption{}
        \label{fig:ring:ar}
    \end{subfigure}%
    ~%
    \begin{subfigure}[t]{0.24\textwidth}
        \includegraphics[width=\textwidth]{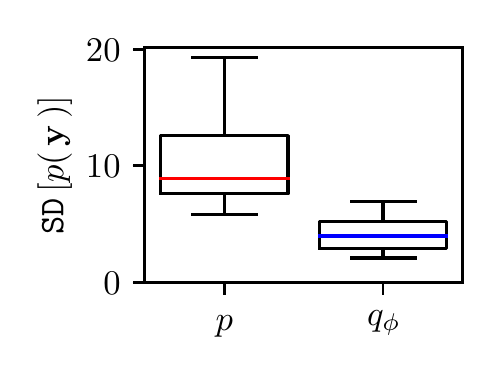}
        \vspace*{-0.7cm}
        \caption{}
        \label{fig:ring:smc:var}
    \end{subfigure}
\vspace*{-0.2cm}
\caption{Results for the annulus problem introduced in Section \ref{sec:annulus}, where the acceptable region of perturbations is inside the black dashed band.
\ref{fig:ring:space} shows in blue the learned state-dependent proposal distribution over velocity (for a state at rest) is the well-approximating the original proposal (shown in red) \emph{inside} the acceptable region, with minimal mall in the invalid region, all but eliminating rejection as shown in \ref{fig:ring:ar}.
\ref{fig:ring:smc:var} shows the reduction in the variance of the evidence by using $q_{\phi}$.
We compute the variance using $100$ independent SMC sweeps, each using $100$ particles, and compare across $100$ datasets.
}
\label{fig:ring}
\end{figure}

The results of this experiment are shown in Figure~\ref{fig:ring}.
The interior of the black dashed lines in Figure~\ref{fig:ring:space} indicates the permissible $\dot{x}$-$\dot{y}$ perturbation, for the given position and zero velocity, where we have centered each distribution on the current position for ease of visual inspection.
Red contours indicate the original density $p(\mathbf{z}_t | \mathbf{x}_{t-1})$, and blue contours indicate the learned density $q_{\phi}(\mathbf{z}_t | \mathbf{x}_{t-1})$.
The fraction of the probability mass outside the black dashed region is the expected rejection rate.
Figure~\ref{fig:ring:ar} shows the rejection rate drops from approximately $75\%$ under the original model to approximately $4\%$ using a trained $q_{\phi}$.

\begin{figure}[h!]
    \centering
    
    \begin{subfigure}[t]{0.48\textwidth}
        \begin{subfigure}[t]{\textwidth}
            \includegraphics[width=0.22\textwidth]{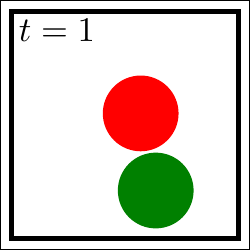}%
            ~
            \includegraphics[width=0.22\textwidth]{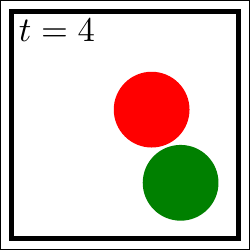}%
            ~
            \includegraphics[width=0.22\textwidth]{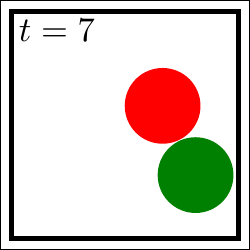}%
            ~
            \includegraphics[width=0.22\textwidth]{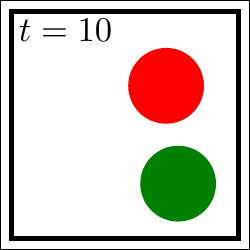}%
        \end{subfigure}
        \vspace*{-0.5cm}
        \caption{}
        \label{fig:balls:trajectory}
    \end{subfigure}%
    
    \begin{subfigure}[t]{0.48\textwidth}
        \includegraphics[width=\textwidth]{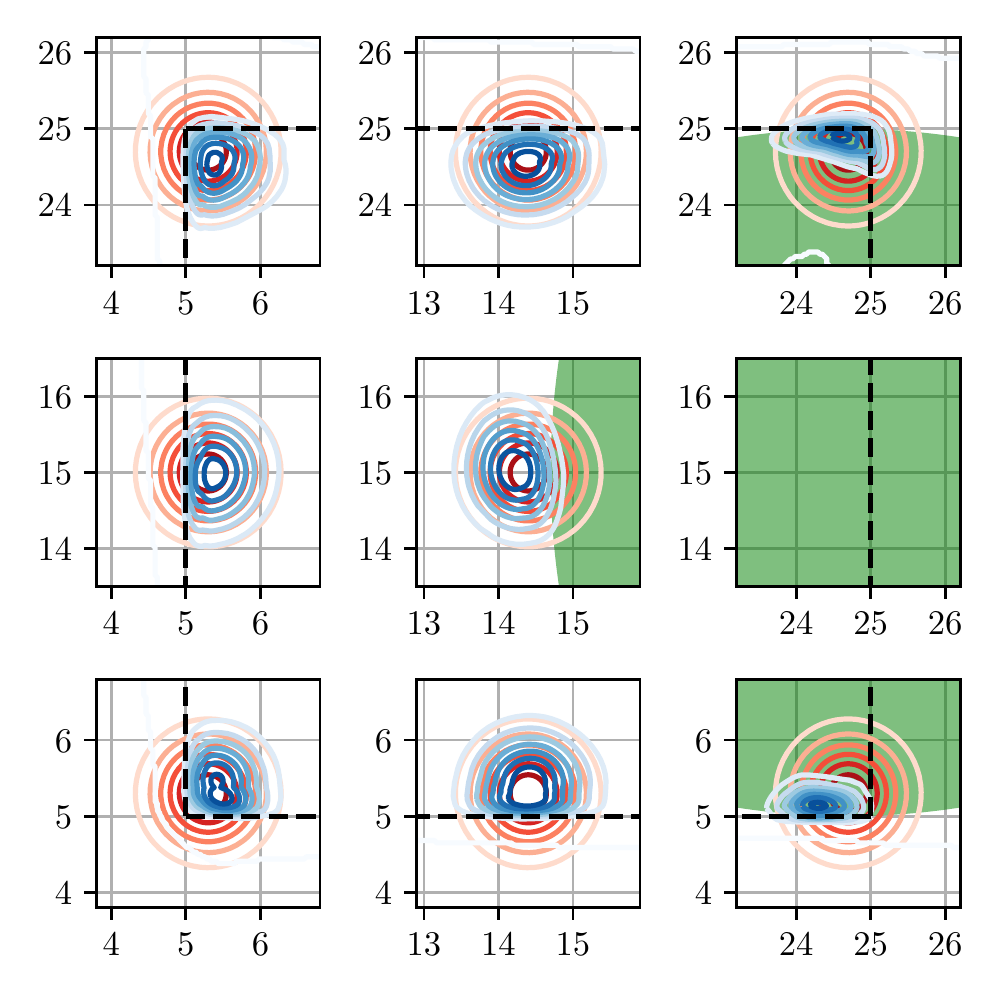}
        \vspace*{-0.7cm}
        \caption{}
        \label{fig:balls:space}
    \end{subfigure}%
    
    \vspace*{0.1cm}
    \begin{subfigure}[t]{0.45\textwidth}
        \begin{overpic}[width=\textwidth]{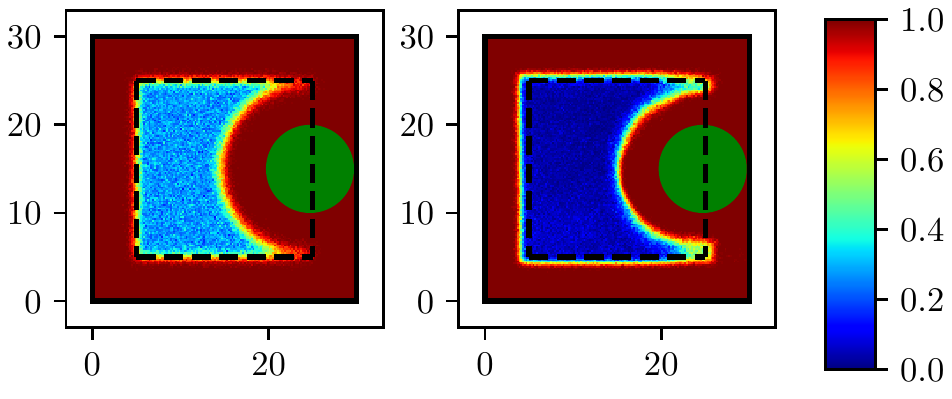}
        \put (4, 4) {$p$}
        \put (44, 4) {$q_{\phi}$}
        \end{overpic}
        \vspace*{-0.5cm}
        \caption{}
        \label{fig:balls:rr}
    \end{subfigure}%
    \vspace*{-0.2cm}
    \caption{Results of the bouncing balls experiment introduced in Section \ref{sec:experiments:bb}, with two radius five, unit mass balls in an enclosure of size $30$.
    \ref{fig:balls:trajectory} shows an example trajectory of the system.
    \ref{fig:balls:space} shows, in red, the proposal distribution over the perturbation to the position of the first ball specified in the model, and the learned proposal in blue. The edge of the permissible region of the enclosure is shown as a black dashed line. The second ball is fixed at $\left[ 25, 15\right]$, and the induced invalid region shaded green. The flow has learned to deflect away from the disallowed regions.
    \ref{fig:balls:rr} shows the rejection rate as a function of the position of the first ball, with the second ball in the position shown. The trained proposal (right) has all but eliminated rejection in the permissible space compared to the a-priori specified proposal (left).
    The rejection rate under $p$ is high in the interior as the second ball may also leave the enclosure, whereas $q_{\phi}$ has practically eliminated rejection by \emph{jointly} proposing perturbations.}
    \label{fig:balls}
\end{figure}

We then use the learned $q_{\phi}$ as the perturbation proposal in an SMC sweep, where we condition on noisy observations of the $x$-$y$ coordinates.
As we focus on the sample efficiency of the sweep, we fix the number of calls to the simulator in Algorithm \ref{alg:rs} to a single call, instead of proposing and rejecting until acceptance.
Failed particles are then not resampled (with certainty) during the resampling.
This means that each iteration of the SMC makes a fixed number of calls to the simulator, and hence we can compare algorithms under a fixed sample budget.
Figure \ref{fig:ring:smc:var} shows that we recover lower variance evidence approximations for a fixed sample budget by using $q_{\phi}$ instead of $p$.
A paired t-test evaluating the difference in variance returns a p-value of less than $0.0001$, indicating a strong statistical difference between the performance under $p$ and $q_{\phi}$, confirming that using $q_{\phi}$ increases the fidelity of inference for a fixed sample budget.

\subsection{Bouncing Balls}
\label{sec:experiments:bb}
Our second example uses a simulator of balls bouncing elastically, as shown in Figure \ref{fig:balls:trajectory}.
We model the position and velocity of each ball, such that the dimensionality of the state vector, $\mathbf{x}_t$, is four times the number of balls.
We add a small amount of Gaussian noise at each iteration to the position and velocity of each ball.
This perturbation induces the possibility that two balls overlap, or, a ball intersects with the wall, representing an invalid physical configuration and results in simulator failure.
We note that here, we are conditioning on the state of \emph{all} balls simultaneously, and proposing the perturbation to the state \emph{jointly}.

Figure \ref{fig:balls:space} shows the distribution over position perturbation of a single ball, conditioned on the other ball being stationary.
Blue contours show the estimated distribution over accepted perturbations learned by autoregressive flow.
Figure \ref{fig:balls:rr} shows the rejection rate under $p$ and $q_{\phi}$ as a function of the position of the first ball, with the second ball fixed in the position shown, showing that rejection has been all but eliminated.
We again see a reduction in the variance of the evidence approximation computed by a particle filter when using $q_{\phi}$ instead of $p$ (figure in the supplementary materials).

\subsection{MuJoCo}
\label{sec:experiments:tosser}
\begin{figure}[t]
    \centering
    
    \begin{subfigure}[t]{0.48\textwidth}
        \centering
        \includegraphics[width=0.24\textwidth]{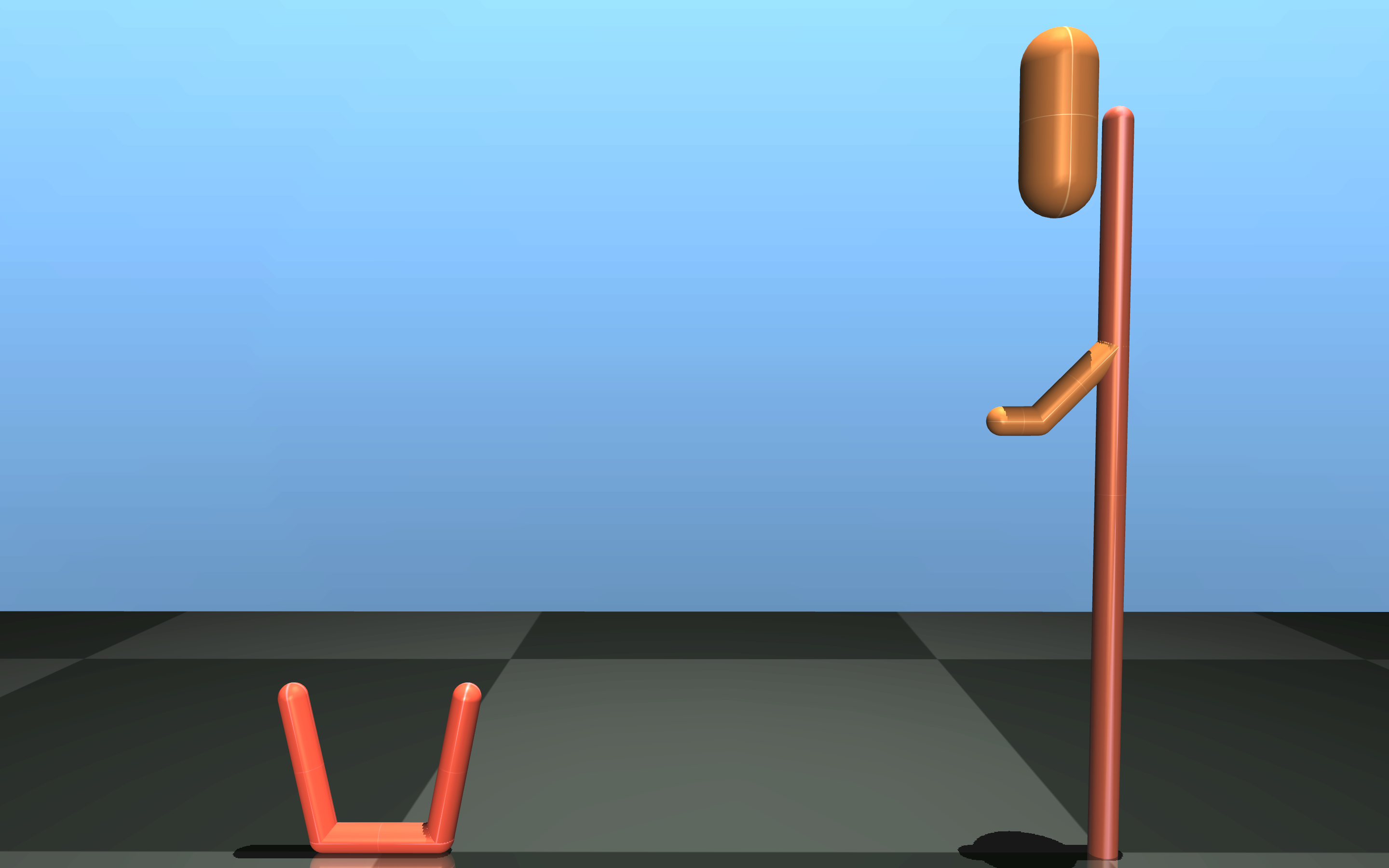}%
        ~%
        \includegraphics[width=0.24\textwidth]{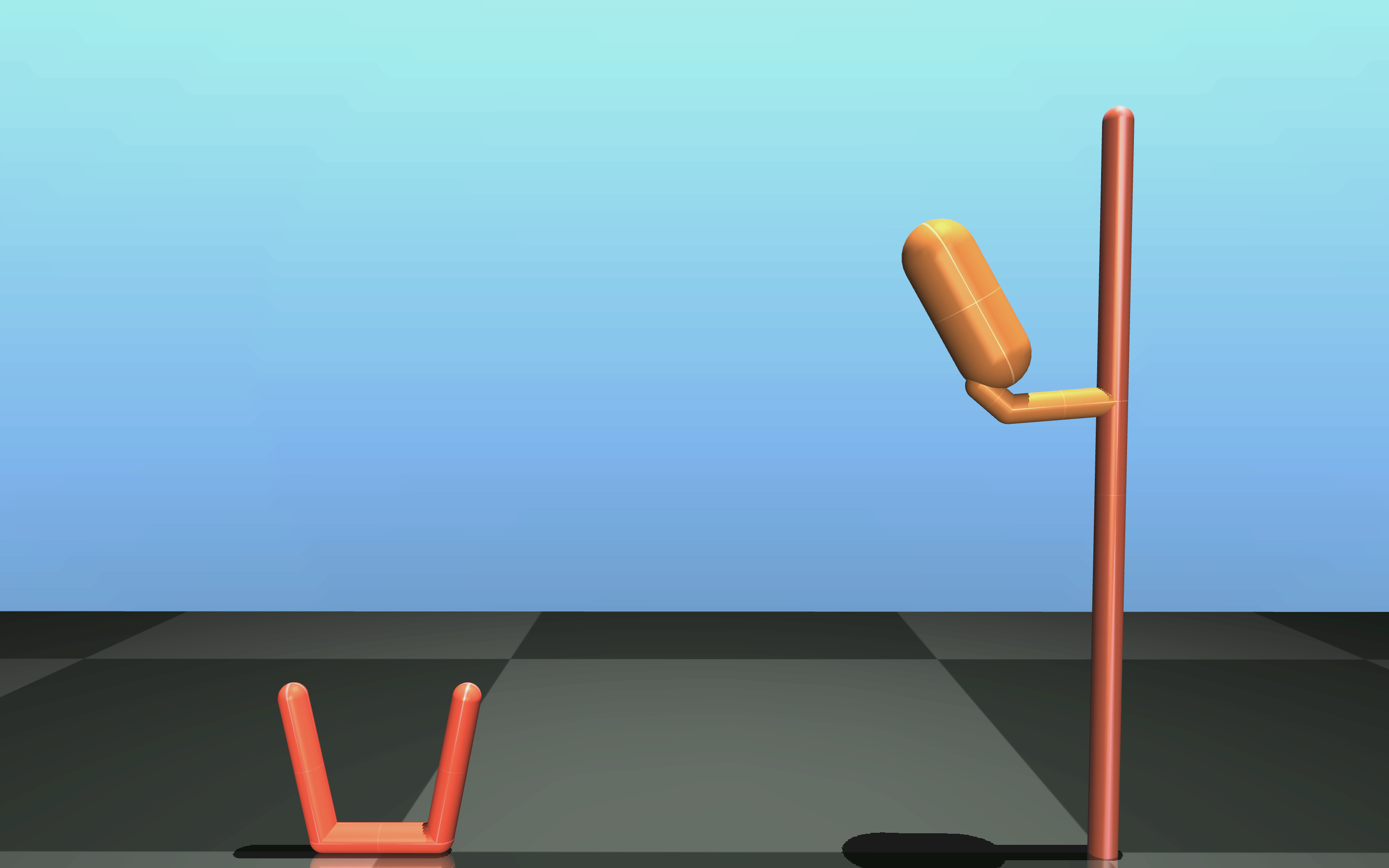}%
        ~%
        \includegraphics[width=0.24\textwidth]{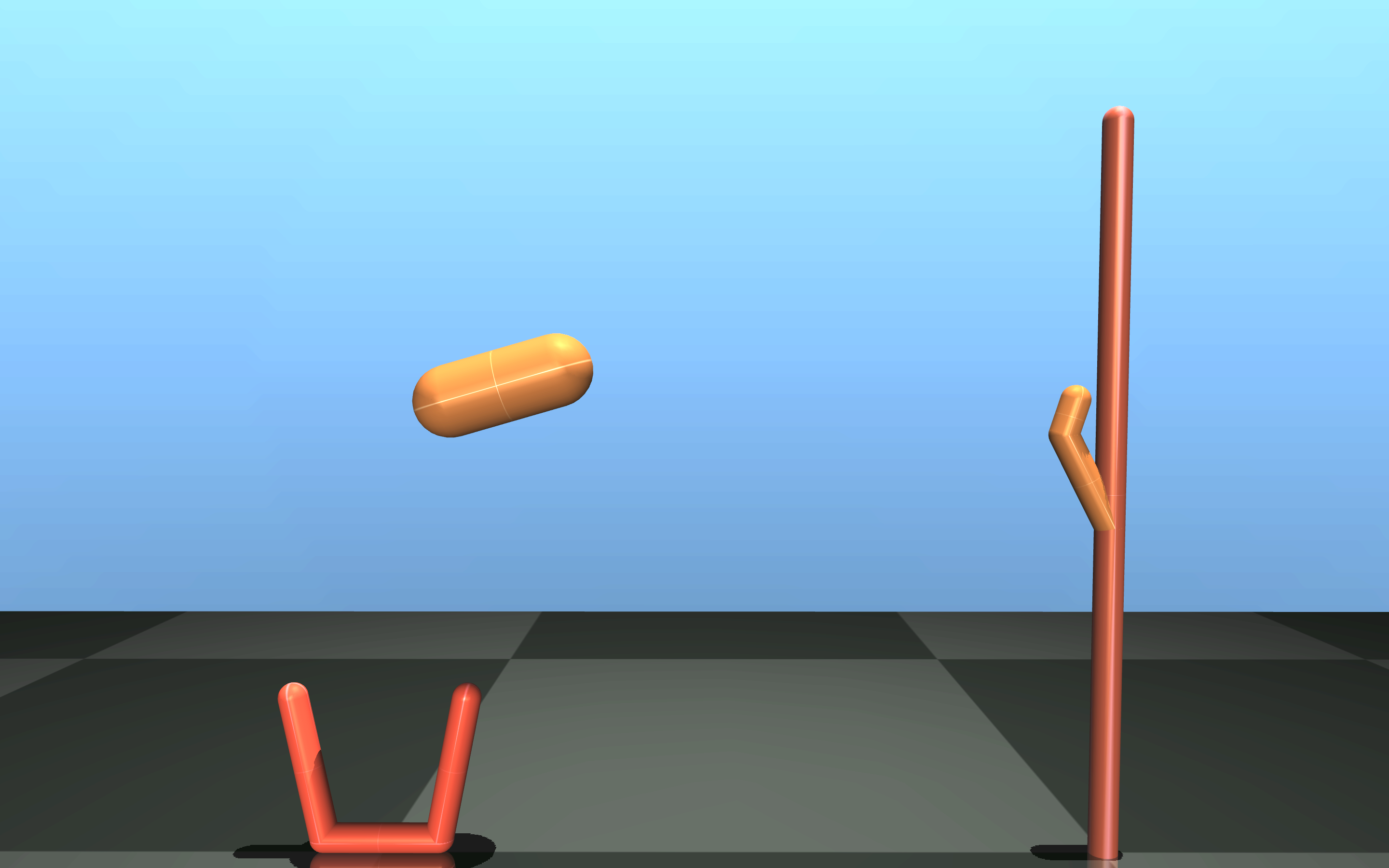}%
        ~%
        \includegraphics[width=0.24\textwidth]{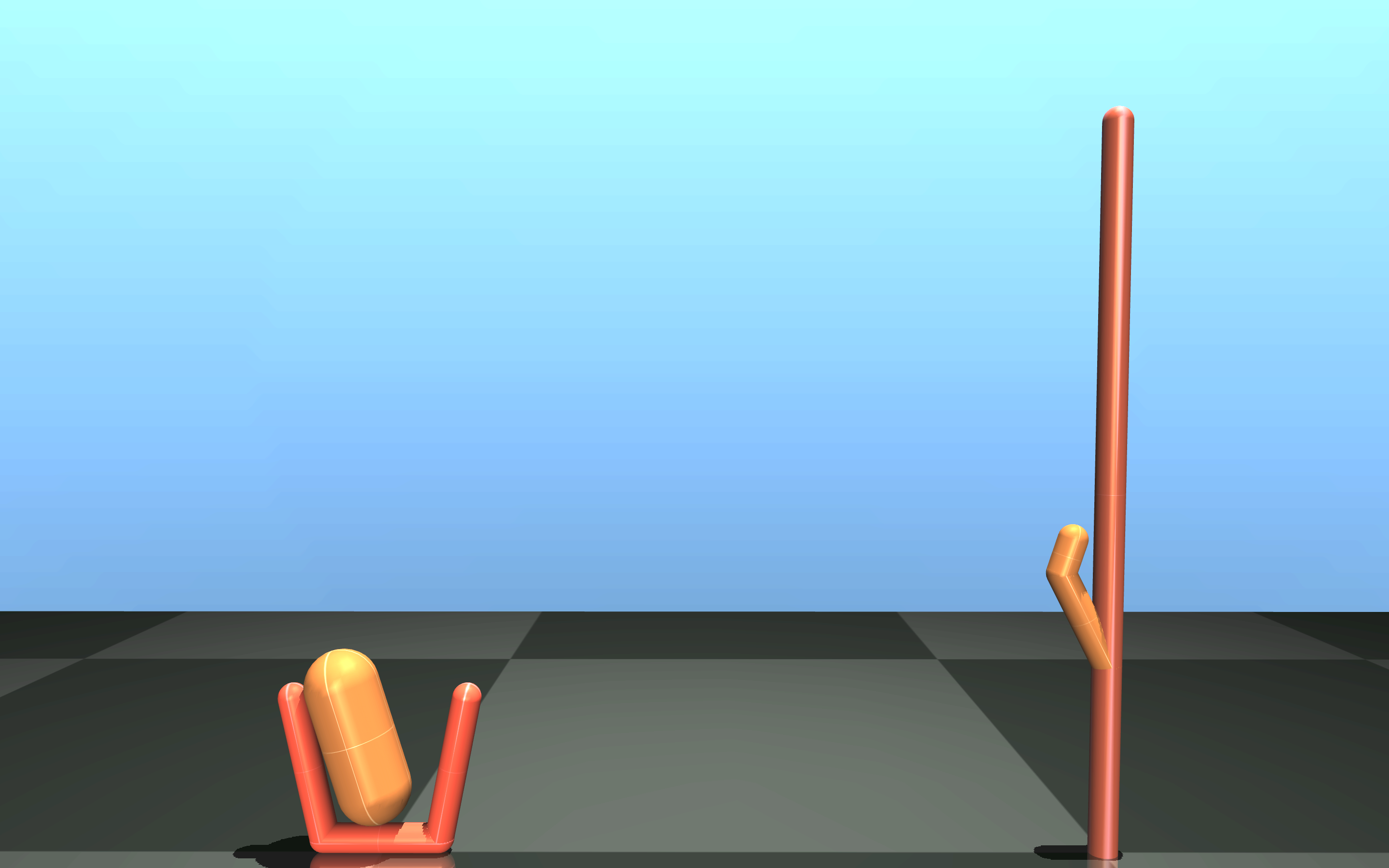}
        \vspace{-0.5cm}
        \caption{ }
        \label{fig:tosser:im}
    \end{subfigure}
    
    \begin{subfigure}[t]{0.24\textwidth}
        \centering
        \includegraphics[width=\textwidth]{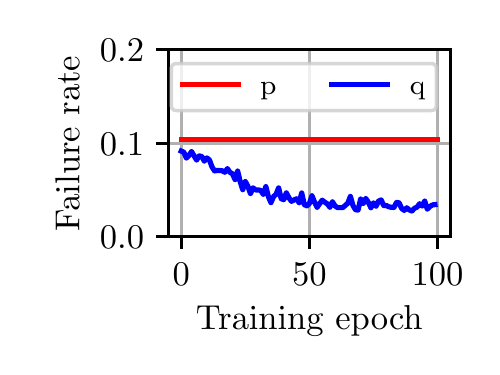}
        \vspace{-0.8cm}
        \caption{ }
        \label{fig:tosser:ar}
    \end{subfigure}%
    ~ 
    \begin{subfigure}[t]{0.24\textwidth}
        \centering
        \includegraphics[width=\textwidth]{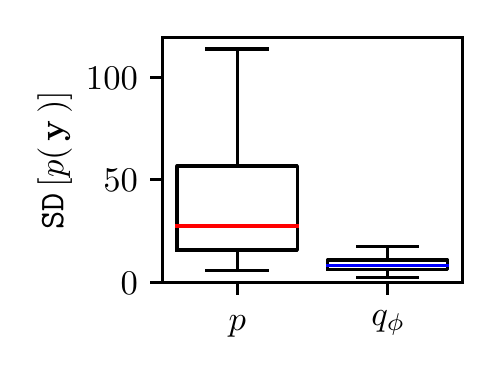}
        \vspace{-0.8cm}
        \caption{ }
        \label{fig:tosser:smc:var}
    \end{subfigure}
    
    \begin{subfigure}[t]{0.48\textwidth}
        \centering
        \includegraphics[width=\textwidth]{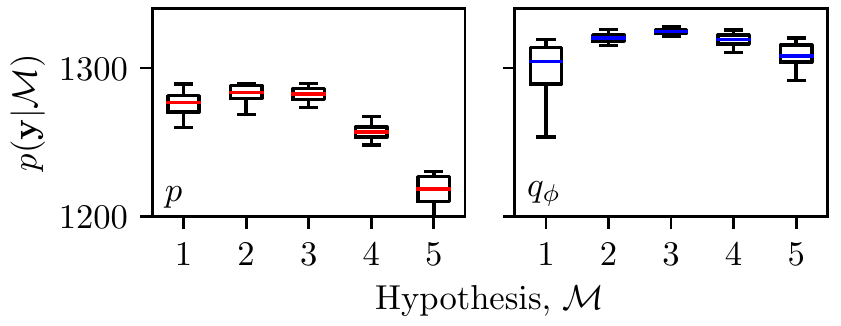}
        \caption{ }
    \label{fig:tosser_hyp}
    \end{subfigure}
    \vspace{-0.2cm}
    \caption{Results of the ``tosser'' experiment introduced in Section \ref{sec:experiments:tosser}.
    \ref{fig:tosser:im} shows the evolution of state over time.
    \ref{fig:tosser:ar} shows the AF we learn markedly reduces the number of rejections.
    \ref{fig:tosser:smc:var} shows the results of performing SMC using the \emph{a priori} specified proposal and our learned autoregressive flow. 
    The autoregressive flow attains a much lower variance estimate, with a p-value of less than $0.0001$ in a paired t-test, indicating a strong statistical difference in performance.
    \ref{fig:tosser_hyp} shows the results of performing hypothesis testing, where hypothesis $3$ is correct, under a uniform prior over hypothesis. 
    The incorrect hypothesis is selected using $p$, while using $q_{\phi}$ the correct hypothesis is selected, with a statistically significant confidence.
    }
    \label{fig:tosser}
\end{figure}

We now apply our method to the popular robotics simulator MuJoCo~\citep{todorov2012mujoco}, specifically using the built-in example ``tosser,'' where a capsule is ``tossed'' by an actuator into a bucket, shown in Figure \ref{fig:tosser:im}.
Tosser displays ``choatic'' aspects, as minor changes in the position of the object results in large changes in the trajectories achieved by the simulator.

MuJoCo allows some overlap between the objects to simulate contact dynamics. 
This is an example of model misspecification borne out of the requirements of reasonably writing a simulator.
We therefore place a hard limit on the amount objects are allowed to overlap.
This is an example of a user-specified constraint that requires the simulator to be run to evaluate.
We add Gaussian distributed noise to the position and velocity of the capsule.

Figure \ref{fig:tosser} shows the results of this experiment.
The capsule is mostly in free space resulting in an average rejection rate under $p$ of $10\%$.
Figure \ref{fig:tosser:ar} shows that the autoregressive flow learns a proposal with a lower rejection rate, reaching $3\%$ rejection.
However these rejections are concentrated in the critical regions of state-space, where chaotic behavior occurs, and so this reduction yields an large reduction in the variance of the evidence approximation, as shown in Figure \ref{fig:tosser:smc:var}.

\begin{figure}[t]
\centering
    \begin{subfigure}[t]{0.24\textwidth}
        \includegraphics[width=\textwidth]{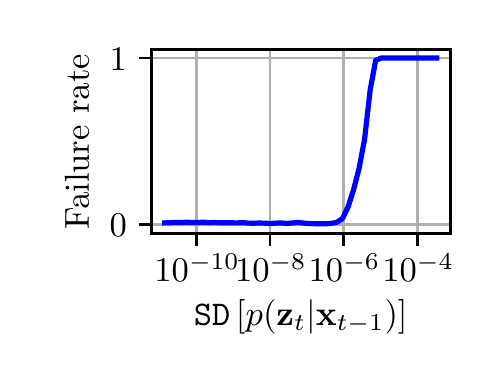}
        \vspace{-0.8cm}
        \caption{}
        \label{fig:wormsim:bot_rate}
    \end{subfigure}%
    ~%
    \begin{subfigure}[t]{0.24\textwidth}
        \includegraphics[width=\textwidth]{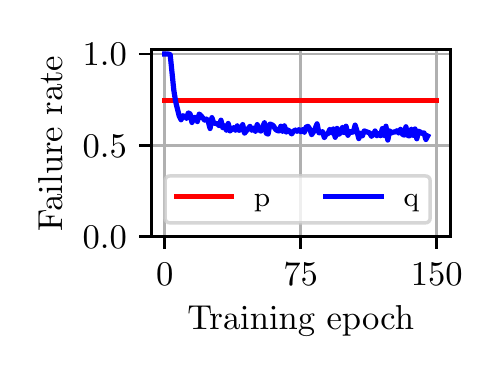}
        \vspace{-0.8cm}
        \caption{}
        \label{fig:wormsim:bot_nf_rate}
    \end{subfigure}
\vspace{-0.5cm}
\caption{Results from the WormSim example introduced in Section \ref{sec:sub:wormsim}.
\ref{fig:wormsim:bot_rate} shows the rate at which the simulator fails increases sharply as a function of the standard deviation of the applied perturbation. 
\ref{fig:wormsim:bot_nf_rate} shows the reduction in rejections during training.
}
\label{fig:bot}
\end{figure}

We conclude this example by evaluating our method on hypothesis testing using pseudo-marginal evidence estimates.
The results for this are shown in Figure \ref{fig:tosser_hyp}.
We test $5$ different hypothesis of the mass of the capsule.
Using $p$ results in higher variance evidence approximations than when $q_{\phi}$ is used. 
Additionally, under $p$ the wrong model is selected ($2$ instead of $3$), although with low significance ($p=0.125$), while using $q_{\phi}$ selects the correct hypothesis with $p=0.0127$.
For this experiment we note that $q_{\phi}$ was trained on a single value of mass, and that this ``training mass'' was different to the ``testing mass.''
We believe this contributes to the increased variance in hypothesis $1$, which is very light compared to the training mass.
Training a $q_{\phi}$ with a further level of amortization over different mass values would further increase the fidelity of the model selection.
This is intimately linked with the larger project of jointly learning the model, and so we defer investigation to future works.

\subsection{Neuroscience Simulator}
\label{sec:sub:wormsim}
We conclude by applying our algorithm to a simulator for the widely studied \emph{Caenorhabditis elegans} roundworm.
WormSim, presented by \citet{boyle2012gait}, is a simulator of the locomotion of the worm, using a $510$ dimensional state representation.
We apply perturbations to a $98$ dimensional subspace defining the physical position of the worm, while conditioning on the full $510$ dimensional state vector.
The expected rate of failure increases sharply as a function of the scale of the perturbation applied, as shown in Figure \ref{fig:wormsim:bot_rate}, as the integrator used in WormSim is unable to integrate highly perturbed states.

The rejection rate during training is shown in Figure \ref{fig:wormsim:bot_nf_rate}.
We are able to learn an autoregressive flow with lower rejection rates, reaching approximately $53\%$ rejection, when $p$ has approximately $75\%$ rejection.
Although the rejection rate is higher than ultimately desired, we include this example as a demonstration of how rejections occur in simulators through integrator failure.
We believe larger flows with regularized parameters can reduce the rejection rate further.

\section{Conclusion}
In this paper we have tackled reducing simulator failures caused by naively perturbing the input state.
We achieve this by showing that stochastically perturbed simulators define a rejection sampler with a well defined target distribution and learning a conditional autoregressive flow to estimate the state-dependent proposal distribution conditioned on acceptance.
We then show that using this learned proposal reduces the variance of inference results, with applications for Bayesian model selection.
We believe this work has readily transferable practical contributions to not just the machine learning community, but the wider scientific community where such naively modified simulation platforms are widely deployed.
As part of the experiments we present, we identify an extension: introducing an additional level of amortization over static simulation parameters.
This extension builds towards our larger research vision of building toolchains for efficient inference and learning in brittle simulators.
Further development will facilitate efficient gradient-based model learning in these brittle simulators.

\newpage

\section{Acknowledgements}
Andrew Warrington is supported under the Shilston Scholarship, awarded by Keble College and the Department of Engineering Science, University of Oxford. 
Saeid Naderiparizi and Frank Wood are supported under the Natural Sciences and Engineering Research Council of Canada (NSERC), the Canada CIFAR AI Chairs Program, Compute Canada, Intel, Composites Research Network (CRN) and DARPA under its D3M and LWLL programs.

\bibliographystyle{abbrvnat}
\bibliography{main}

\end{document}